\pdfoutput=1
\documentclass[letterpaper, 10 pt, conference]{ieeeconf}  

\IEEEoverridecommandlockouts                              

\usepackage{amsmath} 
\usepackage{amssymb}  

\usepackage{amsfonts}
\usepackage{cite}
\usepackage{graphicx}
\usepackage{subcaption}
\usepackage{diagbox}
\usepackage{xargs}    
\usepackage[pdftex,dvipsnames,table]{xcolor}
\definecolor{myred}{rgb}{0.9140625,0.26171875,0.20703125}
\definecolor{mygreen}{rgb}{0.203125,0.65625,0.32421875}
\definecolor{myblue}{rgb}{0.2578125,0.51953125,0.953125}
\definecolor{mygreenbright}{rgb}{0.89453125,1.0,0.89453125}
\definecolor{myredbright}{rgb}{1.0,0.89453125,0.89453125}
\definecolor{myyellowbright}{rgb}{1.0,0.984375,0.91796875}
\usepackage{tikz}
\usetikzlibrary{shapes,decorations.pathreplacing,calc}
\def\checkmark{\tikz\fill[scale=0.2](0,.35) -- (.25,0) -- (1,.7) -- (.25,.15) -- cycle;}
\newcommand{\dotgreenbright}{\raisebox{0.5pt}{\tikz{\node[draw,scale=0.8,circle,fill=mygreenbright](){};}}}
\newcommand{\dotredbright}{\raisebox{0.5pt}{\tikz{\node[draw,scale=0.8,circle,fill=myredbright](){};}}}
\newcommand{\dotyellowbright}{\raisebox{0.5pt}{\tikz{\node[draw,scale=0.8,circle,fill=myyellowbright](){};}}}

\newcommand{\tikzmark}[2][-3pt]{\tikz[remember picture, overlay, baseline=-0.5ex]\node[#1](#2){};}

\tikzset{brace/.style={decorate, decoration={brace}},
 brace mirrored/.style={decorate, decoration={brace,mirror}},
}

\newcounter{brace}
\newcommand{\drawbrace}[3][brace]{%
 \refstepcounter{brace}
 \tikz[remember picture, overlay]\draw[#1] (#2.center)--(#3.center)node[pos=0.5, name=brace-\thebrace]{};
}

\newcommand{\annote}[3][]{%
 \tikz[remember picture, overlay]\node[#1] at (#2) {#3};
}

\usepackage[breaklinks=true,bookmarks=false]{hyperref}
\hypersetup{
    colorlinks=true,
    linkcolor=blue,
    filecolor=magenta,      
    urlcolor=cyan,
}

\usepackage{breakurl}

\usepackage[colorinlistoftodos,prependcaption,textsize=tiny]{todonotes}
\newcommandx{\change}[2][1=]{\todo[linecolor=red,backgroundcolor=red!25,bordercolor=red,#1]{#2}}
\newcommandx{\unsure}[2][1=]{\todo[linecolor=blue,backgroundcolor=blue!25,bordercolor=blue,#1]{#2}}
\newcommandx{\good}[2][1=]{\todo[linecolor=OliveGreen,backgroundcolor=OliveGreen!25,bordercolor=OliveGreen,#1]{#2}}
\newcommandx{\improve}[2][1=]{\todo[linecolor=yellow,backgroundcolor=yellow!25,bordercolor=Plum,#1]{#2}}

\title{\LARGE \bf
3D BAT: A Semi-Automatic, Web-based 3D Annotation Toolbox for Full-Surround, Multi-Modal Data Streams}

\author{Walter Zimmer, Akshay Rangesh and Mohan Trivedi
\thanks{The authors are with the Laboratory for Intelligent and Safe Automobiles,
        University of California, San Diego, CA 92092, USA.
        \newline{\tt\small \{wzimmer, arangesh, mtrivedi\}@ucsd.edu}}
}

\begin{document}



\maketitle
\thispagestyle{empty}
\pagestyle{empty}

\begin{abstract}


In this paper, we focus on obtaining 2D and 3D labels, as well as track IDs for objects on the road with the help of a novel 3D Bounding Box Annotation Toolbox (3D BAT). Our open source, web-based 3D BAT incorporates several smart features to improve usability and efficiency. For instance, this annotation toolbox supports semi-automatic labeling of tracks using interpolation, which is vital for downstream tasks like tracking, motion planning and motion prediction. Moreover, annotations for all camera images are automatically obtained by projecting annotations from 3D space into the image domain. In addition to the raw image and point cloud feeds, a Masterview consisting of the top view (bird's-eye-view), side view and front views is made available to observe objects of interest from different perspectives. Comparisons of our method with other publicly available annotation tools reveal that 3D annotations can be obtained faster and more efficiently by using our toolbox.

\end{abstract}

\section{Introduction}
There is a growing need for broadly available 3D annotation platforms that have a labeling user interface that is intuitive enough for non-domain experts to use. Many existing tools can annotate 2D images, but there is a growing need to also annotate 3D LiDAR data. 

Navigating/labeling in 3D space requires a carefully designed user interface. Annotating large amounts of 3D data also requires a big workforce that may need training. Another challenge is that the resolution and the clarity may be limited, making it hard to differentiate between objects. In this paper, we propose an open source tool that alleviates the need to buy end-to-end ``Training Data as a Service" solutions. Annotators can label their own data very accurately and have full control of the toolbox and their data. Our toolbox also provides functionality to evaluate annotated data (built-in quality control) in order to obtain labels with high quality. Finally, human annotation tasks are very labor-intensive and sometimes require additional human resources. As a remedy, one can even consider using our toolbox to outsource the annotation tasks to Amazon Mechanical Turks (AMTs) - a popular crowd-sourcing platform. This is possible because of the web-based, platform-independent nature of our toolbox, capable of running on most modern browsers, irrespective of the operating system involved.

Recently, applications related to autonomous and partially automated cars have attracted significant attention. For research in these applications, a full-surround multi-modal dataset with 2D and 3D annotations, as well as assigned track IDs would be very useful~\cite{rangesh2018no, deo2018would, rangesh2018ground}. Motivated by these needs, our work focuses on the challenging task of 2D image and 3D point cloud annotation of street scenes. Inspired by the easy usage of 3D modeling tools (e.g. Blender \cite{blenderBlenderOrgHome2019}), we make the user annotate scenes directly in 3D and transfer those annotations back into the image domain. Laser data is first annotated with rough bounding primitives, and then a geometric model is used to transfer these labels into the image space. The advantage of this approach is that the label in 3D is automatically projected into multiple camera views, thus lowering the annotation time considerably. We also include additional features like smart interpolation for users to semi-automatically label sequential data streams. All these features put together make our tool easy to use, accurate and efficient.

\begin{figure}[t]
  \centering
  \includegraphics[width=1.0\linewidth]{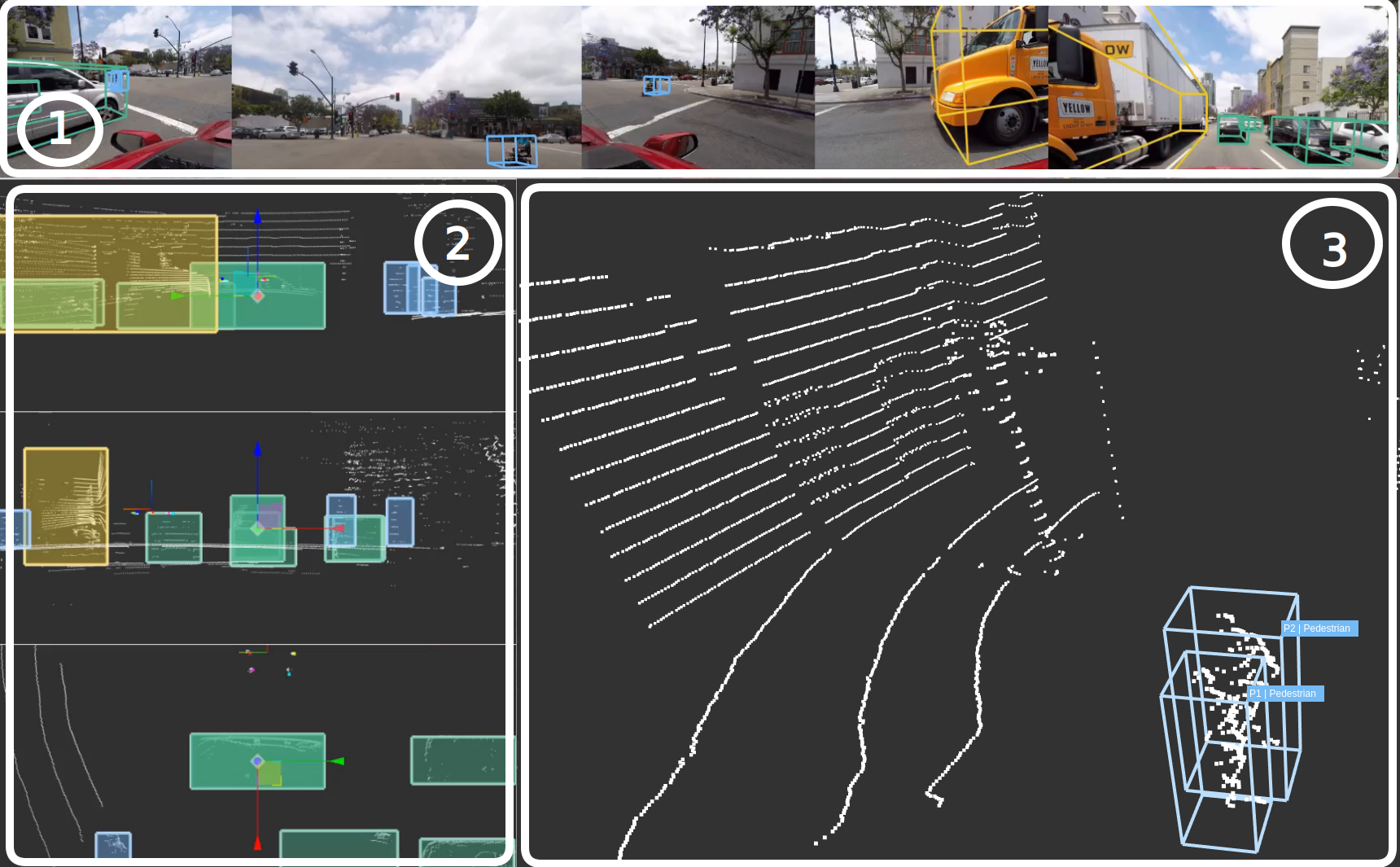}
  \caption{Overview of the annotation toolbox. (1) A horizontal scrollable and vertical resizable panoramic camera image provides a full-surround view. (2) The Masterview that consists of a side view (top), a front view (middle) and a top view (bottom) supports the user during the annotation process. (3) 3D view in that the user can navigate and place annotations.}
  \label{fig:overview}
\end{figure}

The main contributions of this work are as follows: First, we propose a novel annotation system with tools for efficient, accurate 3D localization and tracking of objects using full-surround multi-modal data streams (see Fig.~\ref{fig:overview}). Second, a systematic comparison with 33 annotation tools is carried out to highlight improvements and benefits of our approach (see Table \ref{tbl:comparison}). Finally, we conduct both qualitative and quantitative evaluations of the efficiency and accuracy of our system using four human annotators with diverse background and skills, resulting in more than 13,800 object annotations within one hour of usage. The annotation toolbox is available for public use at: \href{https://github.com/walzimmer/3d-bat}{https://github.com/walzimmer/3d-bat}.

\begin{table*}[t]
  \center
  \captionsetup{justification=centering}
  \caption{Comparison between different annotation tools.\newline \protect\dotgreenbright ~Integrates feature \protect\dotyellowbright  ~Feature only available for 2D images \protect\dotredbright ~Feature not available
  }
  \label{tbl:comparison}
  \begin{tabular}{|l|cccccccccccccc|r|}
    \hline
    \backslashbox{Tool}{Feature}         & F1  & F2  & F3 & F4 & F5 & F6 & F7 & F8 & F9 & F10 & F11 & F12 & F13 & F14 & Score\\
    \hline\hline
    3D BAT (OUR) &  \cellcolor{green!10}\checkmark  &  \cellcolor{green!10}\checkmark  & \cellcolor{green!10}\checkmark  & \cellcolor{green!10}\checkmark  & \cellcolor{green!10}\checkmark  & \cellcolor{green!10}\checkmark  & \cellcolor{green!10}\checkmark  & \cellcolor{green!10}\checkmark  & \cellcolor{green!10}\checkmark  & \cellcolor{green!10}\checkmark   &  \cellcolor{green!10}\checkmark  &  \cellcolor{green!10}\checkmark  &  \cellcolor{green!10}\checkmark &  \cellcolor{green!10}\checkmark  & 14/14\\
    \tikzmark[xshift=-8pt,yshift=1ex]{x}TUBS\cite{plachetkaTUBSRoadUser2018}   &\cellcolor{green!10}\checkmark&\cellcolor{green!10}\checkmark&\cellcolor{green!10}\checkmark&\cellcolor{green!10}\checkmark&\cellcolor{red!10}&\cellcolor{red!10}&\cellcolor{green!10}\checkmark&\cellcolor{red!10}&\cellcolor{green!10}\checkmark&\cellcolor{red!10}   &\cellcolor{red!10}&\cellcolor{red!10}&\cellcolor{red!10}&  \cellcolor{green!10}\checkmark& 7/14\\
    VAST\cite{bergerVASTVolumeAnnotation2018} &\cellcolor{red!10}&\cellcolor{green!10}\checkmark&\cellcolor{green!10}\checkmark&\cellcolor{red!10}&\cellcolor{red!10}&\cellcolor{green!10}\checkmark&\cellcolor{red!10}&\cellcolor{red!10}&\cellcolor{green!10}\checkmark&\cellcolor{red!10}&\cellcolor{red!10}&\cellcolor{red!10}&\cellcolor{green!10}\checkmark&  \cellcolor{green!10}\checkmark &6.0/14\\
    CVAT\cite{vondrickEfficientlyScalingCrowdsourced2013}     &\cellcolor{red!10}& \cellcolor{green!10}\checkmark &\cellcolor{red!10}&\cellcolor{red!10}&\cellcolor{red!10}&\cellcolor{red!10}&\cellcolor{red!10}&\cellcolor{red!10}&\cellcolor{yellow!10}(\checkmark)&\cellcolor{green!10}\checkmark&\cellcolor{green!10}\checkmark&\cellcolor{red!10}&\cellcolor{green!10}\checkmark &  \cellcolor{green!10}\checkmark & 5.5/14\\
    LabelMe\cite{russellLabelMeDatabaseWebBased2008} &\cellcolor{red!10}&\cellcolor{green!10}\checkmark&\cellcolor{red!10}&\cellcolor{red!10}&\cellcolor{red!10}&\cellcolor{red!10}&\cellcolor{red!10}&\cellcolor{red!10}&\cellcolor{yellow!10}(\checkmark)&\cellcolor{green!10}\checkmark&\cellcolor{green!10}\checkmark&\cellcolor{red!10}&\cellcolor{green!10}\checkmark &  \cellcolor{green!10}\checkmark& 5.5/14\\
    CLEAN\cite{kuoImpactAutomaticPreannotation2018} &\cellcolor{red!10}&\cellcolor{green!10}\checkmark&\cellcolor{red!10}&\cellcolor{red!10}&\cellcolor{red!10}&\cellcolor{red!10}&\cellcolor{red!10}&\cellcolor{red!10}&\cellcolor{yellow!10}(\checkmark)&\cellcolor{green!10}\checkmark&\cellcolor{green!10}\checkmark&\cellcolor{red!10}&\cellcolor{green!10}\checkmark &  \cellcolor{green!10}\checkmark & 5.5/14\\
    BRAT\cite{stenetorpBratWebbasedTool2012} &\cellcolor{red!10}&\cellcolor{green!10}\checkmark&\cellcolor{red!10}&\cellcolor{red!10}&\cellcolor{red!10}&\cellcolor{red!10}&\cellcolor{red!10}&\cellcolor{red!10}&\cellcolor{yellow!10}(\checkmark)&\cellcolor{green!10}\checkmark&\cellcolor{red!10}&\cellcolor{red!10}&\cellcolor{green!10}\checkmark &  \cellcolor{red!10}& 3.5/14\\
    WebAnno\cite{decastilhoWebbasedToolIntegrated2016}      &\cellcolor{red!10}&\cellcolor{red!10}&\cellcolor{red!10}&\cellcolor{red!10}&\cellcolor{red!10}&\cellcolor{red!10}&\cellcolor{red!10}&\cellcolor{red!10}&\cellcolor{yellow!10}(\checkmark)&\cellcolor{green!10}\checkmark&\cellcolor{red!10}&\cellcolor{red!10}&\cellcolor{green!10}\checkmark &  \cellcolor{green!10}\checkmark& 3.5/14\\
    \tikzmark[xshift=-8pt,yshift=-1ex]{y}Image Tagger\cite{fiedlerImageTaggerOpenSource2018} &\cellcolor{red!10}&\cellcolor{red!10}&\cellcolor{red!10}&\cellcolor{red!10}&\cellcolor{red!10}&\cellcolor{red!10}&\cellcolor{red!10}&\cellcolor{red!10}&\cellcolor{yellow!10}(\checkmark)&\cellcolor{green!10}\checkmark&\cellcolor{red!10}&\cellcolor{red!10}&\cellcolor{green!10}\checkmark &  \cellcolor{green!10}\checkmark& 3.5/14\\

    \hline\hline
    3D BAT (OUR) &  \cellcolor{green!10}\checkmark  &  \cellcolor{green!10}\checkmark  & \cellcolor{green!10}\checkmark  & \cellcolor{green!10}\checkmark  & \cellcolor{green!10}\checkmark  & \cellcolor{green!10}\checkmark  & \cellcolor{green!10}\checkmark  & \cellcolor{green!10}\checkmark  & \cellcolor{green!10}\checkmark  & \cellcolor{green!10}\checkmark   &  \cellcolor{green!10}\checkmark  &  \cellcolor{green!10}\checkmark  &  \cellcolor{green!10}\checkmark  &  \cellcolor{green!10}\checkmark& 14/14\\
    \tikzmark[xshift=-8pt,yshift=1ex]{u}scale.ai\cite{scaleaiSensorFusion3D2019} &\cellcolor{green!10}\checkmark&\cellcolor{green!10}\checkmark&\cellcolor{green!10}\checkmark&\cellcolor{red!10}&\cellcolor{green!10}\checkmark&\cellcolor{green!10}\checkmark&\cellcolor{green!10}\checkmark&\cellcolor{red!10}&\cellcolor{green!10}\checkmark&\cellcolor{green!10}\checkmark&\cellcolor{red!10}&\cellcolor{red!10}&\cellcolor{green!10}\checkmark &  \cellcolor{green!10}\checkmark& 10/14\\
    playment.io\cite{playment3DPointCloud2019}  &\cellcolor{red!10}&\cellcolor{green!10}\checkmark&\cellcolor{green!10}\checkmark&\cellcolor{green!10}\checkmark&\cellcolor{green!10}\checkmark&\cellcolor{green!10}\checkmark&\cellcolor{green!10}\checkmark&\cellcolor{red!10}&\cellcolor{green!10}\checkmark&\cellcolor{green!10}\checkmark&\cellcolor{red!10}&\cellcolor{red!10}&\cellcolor{red!10} &  \cellcolor{green!10}\checkmark& 9/14\\
    surfing.ai\cite{xieSemanticInstanceAnnotation2016}   &\cellcolor{green!10}\checkmark&\cellcolor{green!10}\checkmark&\cellcolor{green!10}\checkmark&\cellcolor{red!10}&\cellcolor{red!10}&\cellcolor{green!10}\checkmark&\cellcolor{red!10}&\cellcolor{green!10}\checkmark&\cellcolor{green!10}\checkmark&\cellcolor{green!10}\checkmark   &\cellcolor{red!10}&\cellcolor{red!10}&\cellcolor{green!10}\checkmark &  \cellcolor{green!10}\checkmark& 9/14\\
    dataturks.com\cite{dataturksBestOnlinePlatform2019}&\cellcolor{red!10}& \cellcolor{green!10}\checkmark &\cellcolor{red!10}&\cellcolor{red!10}&\cellcolor{red!10}&\cellcolor{red!10}&\cellcolor{red!10}&\cellcolor{red!10}&\cellcolor{yellow!10}(\checkmark)&\cellcolor{green!10}\checkmark&\cellcolor{green!10}\checkmark&\cellcolor{red!10}&\cellcolor{green!10}\checkmark &  \cellcolor{green!10}\checkmark& 5.5/14\\
    supervise.ly\cite{superviselySuperviselyWebPlatform2019} &\cellcolor{red!10}&\cellcolor{red!10}&\cellcolor{red!10}&\cellcolor{red!10}&\cellcolor{red!10}&\cellcolor{red!10}&\cellcolor{red!10}&\cellcolor{red!10}&\cellcolor{yellow!10}(\checkmark)&\cellcolor{green!10}\checkmark&\cellcolor{green!10}\checkmark&\cellcolor{red!10}&\cellcolor{green!10}\checkmark &  \cellcolor{green!10}\checkmark& 4.5/14\\
    labelbox.com\cite{labelboxLabelboxBestWay2019} &\cellcolor{red!10}&\cellcolor{red!10}&\cellcolor{red!10}&\cellcolor{red!10}&\cellcolor{red!10}&\cellcolor{red!10}&\cellcolor{red!10}&\cellcolor{red!10}&\cellcolor{yellow!10}(\checkmark)&\cellcolor{green!10}\checkmark&\cellcolor{green!10}\checkmark&\cellcolor{red!10}&\cellcolor{green!10}\checkmark &  \cellcolor{green!10}\checkmark& 4.5/14\\
    neurala.com\cite{neuralaBrainBuilderBeta2019}  &\cellcolor{red!10}&\cellcolor{red!10}&\cellcolor{red!10}&\cellcolor{red!10}&\cellcolor{red!10}&\cellcolor{red!10}&\cellcolor{red!10}&\cellcolor{red!10}&\cellcolor{yellow!10}(\checkmark)&\cellcolor{green!10}\checkmark&\cellcolor{green!10}\checkmark&\cellcolor{red!10}&\cellcolor{green!10}\checkmark &  \cellcolor{green!10}\checkmark& 4.5/14\\
    \tikzmark[xshift=-8pt,yshift=-1ex]{v}prodi.gy\cite{prodigyProdigyAnnotationTool2019}     &\cellcolor{red!10}&\cellcolor{red!10}&\cellcolor{red!10}&\cellcolor{red!10}&\cellcolor{red!10}&\cellcolor{red!10}&\cellcolor{red!10}&\cellcolor{red!10}&\cellcolor{yellow!10}(\checkmark)&\cellcolor{green!10}\checkmark&\cellcolor{green!10}\checkmark&\cellcolor{red!10}&\cellcolor{green!10}\checkmark &  \cellcolor{green!10}\checkmark& 4.5/14\\
    \hline
  \end{tabular}
  \drawbrace[brace mirrored, thick]{x}{y}
  \annote[left]{brace-1}{\scriptsize open source~~}
  \drawbrace[brace mirrored, thick]{u}{v}
  \annote[left]{brace-2}{\scriptsize commercial~~}
\end{table*}

\begin{table*}
    \centering
  \begin{tabular}{llp{1cm}ll}
    F1 & Full-surround annotations & &  F8 & 3D transform controls\\
    F2 & Semi-automatic labeling & &F9 & 2D and 3D annotations\\
    F3 & 3D to 2D label transfer (projections) & &F10 & Web-based (online accessible \& platform ind.)\\
    F4 & Automatic tracking && F11 & Redo/undo functionality\\
    F5 & Masterview (side, front, top and 3D view) && F12 & Keyboard only annotation mode\\
    F6 & Navigation in 3D && F13 & Auto save function\\
    F7 & Auto ground detection && F14 & Review annotations \\
      \end{tabular}
\end{table*}

\section{Related Studies}
In this section, we provide a comprehensive comparison between a variety of popular annotation tools. Although 33 annotation tools were evaluated in total, only 16 of the highest scoring ones are shown in Table \ref{tbl:comparison} for the sake of brevity. More than half of the evaluated tools are open source. The comparison is split into two tables to separate the open source tools from the commercial tools. 

Looking at the open source tools, only the \texttt{TUBS}\cite{plachetkaTUBSRoadUser2018} annotation tool comes close to our toolbox. It provides full-surround annotations, semi-automatic labeling and annotations are projected to all cameras. Besides an automatic tracking functionality, the \texttt{TUBS} labeling tool has also possibility to detect and remove the ground to better locate road users on the ground. 
Our tool was also compared to commercial tools and achieves the highest ranking, given the features that are listed in Table \ref{tbl:comparison}. Among the commercial options, three tools come close to our annotation tool: \texttt{scale.ai}\cite{scaleaiSensorFusion3D2019}, \texttt{playment.io}\cite{playment3DPointCloud2019} and \texttt{surftech.ai}\cite{xieSemanticInstanceAnnotation2016}. All of them provide automatic or semi-automatic 3D annotation techniques, a 3D to 2D label transfer as well as the possibility to navigate in 3D space. Most of the annotation tools listed in the table are web-based, offer a redo/undo functionality and automatically save changes that are done by the user. However, none of the tools support keyboard only annotations, which is a very efficient way to create labels. Full-surround annotations are only offered by \texttt{scale.ai}, \texttt{surfing.ai} and the \texttt{TUBS} annotation tool, whereas only \texttt{playment.io} and the \texttt{TUBS} annotation tool support the annotation of tracks. Compared to annotating individual images, video sequences offer the advantage of temporal coherence between adjacent frames, thereby making it possible to use propagation techniques to transfer labels from the current frame to subsequent frames. Note that two open source tools (\texttt{TUBS} and \texttt{VAST}) and only three commercial tools (\texttt{scale.ai}, \texttt{playment.io} and \texttt{surftech.ai}) support the annotation of 3D objects. All other listed tools provide only the functionality to label in image space, hence only half a point (yellow checkmark) was added to the final score. 

To evaluate our tool in realistic use cases, we consider different publicly available real world datasets. Compared to the popular \texttt{KITTI} dataset which only has camera images that were obtained by two stereo cameras pointing forward, the \texttt{LISA-T}\cite{rangeshExploringSituationalAwareness2018} test vehicle has six full-surround cameras, thus more data has to be annotated per timestamp. We also consider the newly released \texttt{NuScenes} dataset~\cite{caesarDevkitNuScenesDataset2019} comprising of a similar full-surround sensor configuration. Since this dataset consists of vast amounts of labelled sequences, we use it to carry out quantitative evaluations of our tool. 

\section{System Architecture}
In this section, the proposed methodology as well as the implemented features will be described in detail. The goal was to develop a limited and simple interface, since this leads to superior annotation experience. The 3D annotation toolbox is based on \texttt{WebGL} (see Fig. \ref{fig:overview}) to allow collaborative annotating. The toolbox was designed to annotate one object at a time because it is more efficient and strongly preferred by the workers. 

First, all available camera images are displayed at once (full-surround view, see Fig. \ref{fig:overview}) so that objects that are covered by multiple cameras can always be seen by the user. Without this feature, the user would have to switch between all camera images to find an object, leading to inefficiency. 

\begin{figure}[t]
\centering
  \begin{subfigure}[t]{\linewidth}
    \includegraphics[height=4.2cm]{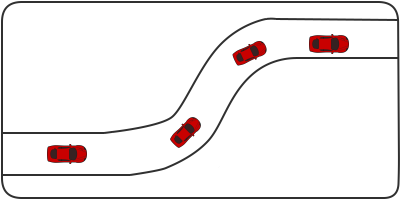}
    \caption{Creating control points for interpolation.}
    \label{fig:interpolation_illustration}
  \end{subfigure}
  \begin{subfigure}[t]{\linewidth}
    \includegraphics[height=4.2cm]{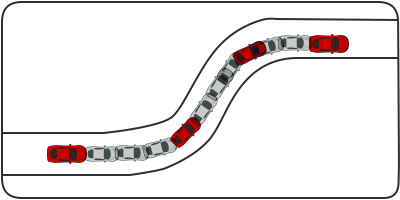}
    \caption{Interpolation of objects. The track of the car is displayed transparent (gray). The interpolation takes also rotation and scaling into account.}
    \label{fig:interpolation_illustration2}
  \end{subfigure}
  \caption{Illustration of object interpolation within a curve.\label{fig:interpolation}}
\end{figure}

Next, an optional semi-automatic labeling method is used to annotate all frames between two specific frames (start and end frame) that are determined by the annotator. After the implementation of this semi-automatic interpolation technique (see Fig. \ref{fig:interpolation_illustration} and \ref{fig:interpolation_illustration2}), the annotation time was seen to drastically decrease. This interpolation is based on a linear model and is to be used primarily for labelling small clips of the data sequence.

Additionally, we use 3D and projective geometry to create a 3D to 2D label transfer option for the user. A 3D to 2D label transfer is very useful to obtain automatically annotated camera images from already labelled 3D point clouds. The annotator places a 3D label in the point cloud which is then projected into all six camera images. This method reduces annotation time and helps the user to place the annotation very accurately into the point cloud, since the projection is updated in real-time.
The projection of 3D annotations into the camera images (3D to 2D label transfer) does not require the user to label each camera image which increases the speed by many orders. An automatic track assignment of objects simplifies the annotation further. Those tracks can later be used for motion planning and motion prediction. Adding the \textit{Masterview} (see Fig. \ref{fig:overview}) which consists of the side, front and bird's-eye-view, the user doesn't have to change into the 3D view anymore to adjust the dimensions of an object. 

\begin{figure}[t]
  \begin{subfigure}[t]{0.32\linewidth}
    \includegraphics[height=4cm]{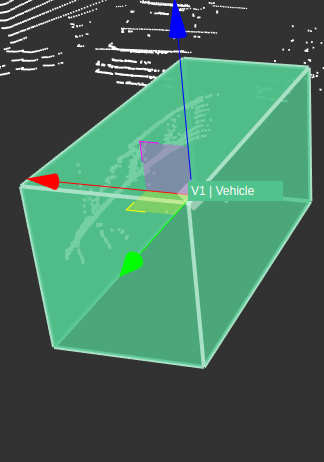}
    \caption{Translation controls.}
    \label{fig:translate}
  \end{subfigure}
  \begin{subfigure}[t]{0.32\linewidth}
    \includegraphics[height=4cm]{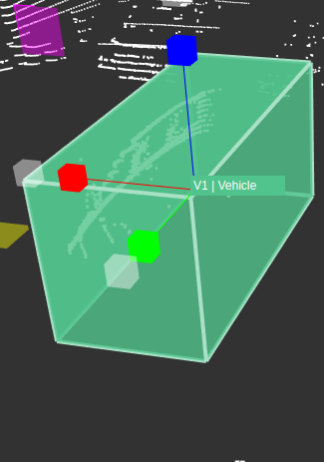}
    \caption{Scale Controls.}
    \label{fig:scale}
  \end{subfigure}
  \begin{subfigure}[t]{0.32\linewidth}
    \includegraphics[height=4cm]{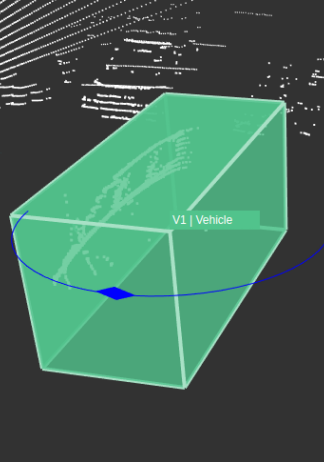}
    \caption{Rotation Controls.}
    \label{fig:rotate}
  \end{subfigure}
  \caption{The user can switch between three transformation modes for which short cuts are provided. Transformation controls can be hidden as well as changed in size. They allow to translate/scale an object along two axis simultaneously.} 
  \label{fig:transformation_controls}
\end{figure}

Once the user selects an object in the scene, transform controls appear on next to it (see Fig. \ref{fig:translate}, \ref{fig:scale} and \ref{fig:rotate}). These controls allow the user to translate, scale and rotate the selected object. The user can interactively correct wrong positions and orientations and change between those three modes using keyboard shortcuts. This transformation can also be done using the keyboard. Shortcuts are provided to switch between the translation, scaling and rotation mode.

\begin{figure*}[t]
  \centering
  \includegraphics[width=0.95\textwidth, height=1.8in]{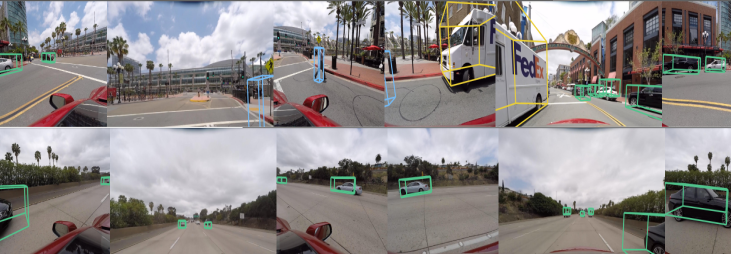}
  \caption{Example annotations of two sequences from the \texttt{LISA-T} dataset.}
  \label{fig:annotations_four_sequences}
\end{figure*}

Furthermore, graphical elements such as buttons and checkboxes were kept at a minimum and keyboard shortcuts are provided for every operation. Having a limited number of choices is shown to save time and remove confusion, as it diminishes the cognitive load of the workers. Fig.~\ref{fig:annotations_four_sequences} shows two example frames that were labeled from the \texttt{LISA-T} dataset.

\section{Experimental Evaluation}
The most common strategy to evaluate annotated data is to compare the obtained annotations with ground truth, this is, annotations from experts. After all features were implemented an experiment was performed in that several user had the task to annotate a sequence of 10 seconds which consists of 300 key frames and another very crowded sequence of 35 key frames. Annotators were first instructed to watch the tutorial videos as well as the raw video to get familiar with the scene and to see where interpolation makes sense. The annotation results were compared against the ground truth that was created by an experienced user (User1) who was familiar with the annotation toolbox. Table \ref{tbl:statistics} shows how much time, mouse clicks and keystrokes the annotation user needed to annotate a sequence. Furthermore the total number of annotated objects is listed as well as the number of objects for each class.

\begin{table}[t]
  \centering
  \caption{Annotation statistics of different users on the \texttt{LISA-T} dataset.}
  \label{tbl:statistics}
  \begin{tabular}{|l|llll|}
    \hline
                             & Ground truth     & User1    & User2    & User3\\
                             \hline\hline
    Time (hh:mm):         &  02:00    & 01:02    & 01:00       & \textbf{00:52}\\
    Clicks:               &  3324     & 1388     & 984         & \textbf{780}\\
    Keystrokes:           &  1032     & \textbf{112}      & 255         & 1663\\
    Annotated obj.:       &  \textbf{6851}     & 1722     & 3078        & 2212\\
    \quad CAR:            &  \textbf{5051}     & 1112     & 2329        & 1481\\
    \quad PEDESTRIAN:     &  \textbf{1500}     & 610      & 749         & 431\\
    \quad MOTORCYCLE:     &  0        & 0        & 0           & 0\\
    \quad BICYCLE:        &  0        & 0        & 0           & 0\\
    \quad TRUCK:          &  300      & 0        & 0           & 300\\
    3D-IoU:               &  -        & 0.073     & \textbf{0.206}        & 0.135\\
    3D-IoU\textgreater0.6:&  -        & 0.058     & \textbf{0.091}        & 0.049\\
    Precision:            &  -        & 0.012     & \textbf{0.013}        & 0.010\\
    Recall:               &  -        & 0.012     & 0.012        & 0.010\\
    $F_1$ score:          &  -        & 0.012     & \textbf{0.091}        & 0.010\\
    \hline
  \end{tabular}
\end{table}

\begin{figure}[t]
  \begin{subfigure}[t]{0.95\linewidth}
    \centering
    \includegraphics[height=4cm]{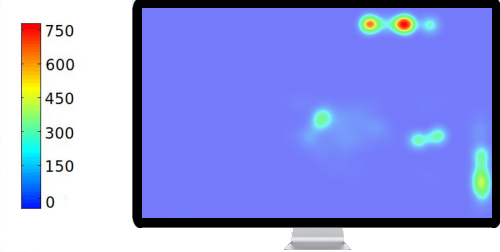}
    \caption{Mouse interactions with the annotation tool. Red areas visualize a high number of clicks.}
    \label{fig:heatmap_mouse}
  \end{subfigure}
  \begin{subfigure}[t]{1.0\linewidth}
    \centering
    \includegraphics[height=2.9cm]{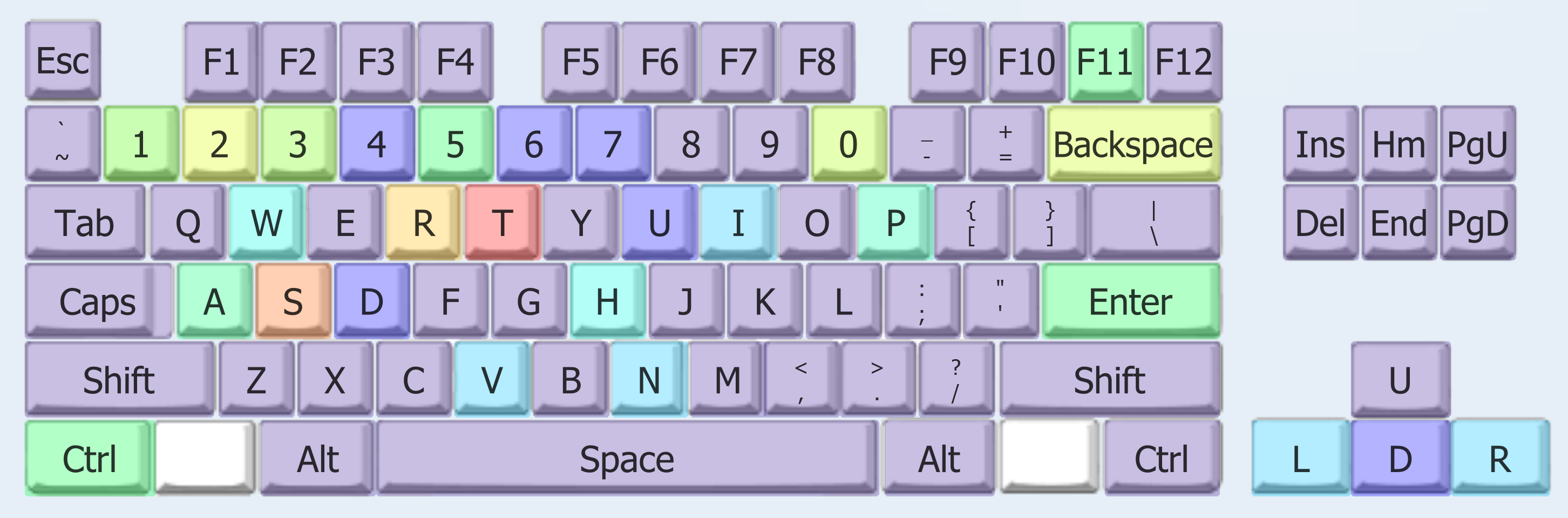}
    \caption{A keyboard heatmap visualizes the keystrokes of all annotators.}
    \label{fig:heatmap_keyboard}
  \end{subfigure}
  \caption{Heatmaps of keyboard and mouse were created using the toolbox WhatPulse\cite{smitWhatPulse2019}.\label{fig:heatmaps}}
\end{figure}

Annotating 1800 camera images and 300 laser scans requires on average about one hour. This results in annotating 30 camera images and 5 laser scans per second. To improve the speed of the annotation process one can evaluate heatmaps of both, mouse and keyboard. Fig. \ref{fig:heatmap_mouse} shows the average heatmap of user interactions with the mouse and Fig. \ref{fig:heatmap_keyboard} visualizes the heatmap of the average keystrokes the annotation users performed during their task.

Evaluating the mouse heatmap one can see that annotators preferred to use the button to navigate to the next frame instead of using a keyboard shortcut which would be faster. A solution would be to provide more instructions to the user a also provide more time to get familiar with the toolbox. The majority of the annotators translated and rescaled the 3D bounding box after they positioned it into the pointcloud. Hence the keyboard key 'T' (to switch into translation mode) and 'S' (to switch into scaling mode) was pressed often. The average intersection over union in 3D (3D-IoU) between the manually created ground truth and the user annotations is 0.138 and 0.066 (considering a boundary of 0.6).

\begin{figure}[t]
  \centering
  \captionsetup{justification=centering}
  \begin{subfigure}[t]{0.49\linewidth}
    \includegraphics[width=1.0\textwidth]{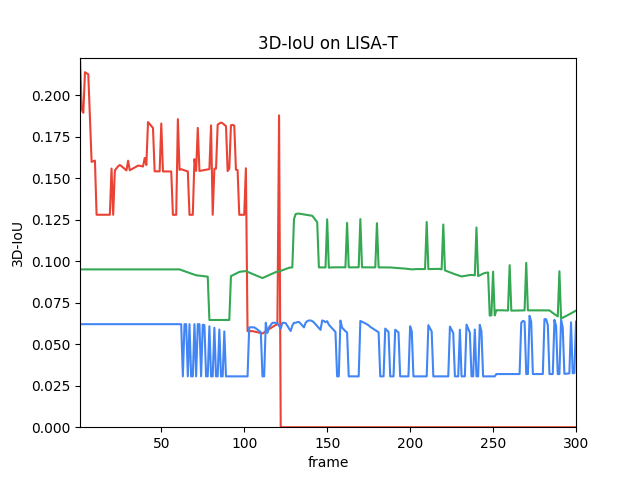}
    \caption{3D-IoU}
    \label{fig:3d_iou_lisa_t}
  \end{subfigure}
  \begin{subfigure}[t]{0.49\linewidth}
    \includegraphics[width=1.0\textwidth]{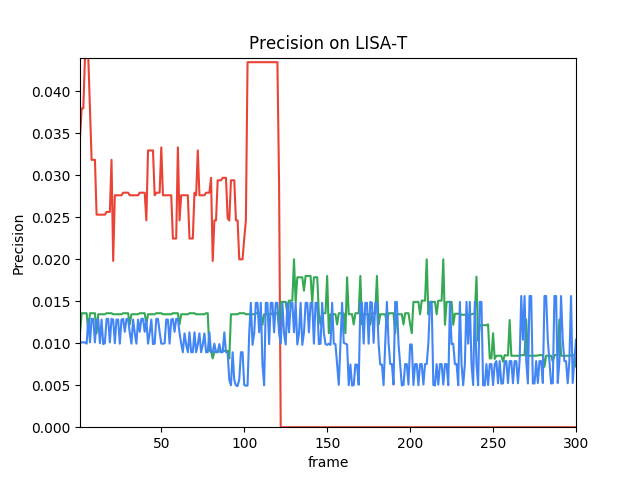}
    \caption{Precision}
    \label{fig:precision_lisa_t}
  \end{subfigure}
  \begin{subfigure}[t]{0.49\linewidth}
    \includegraphics[width=1.0\textwidth]{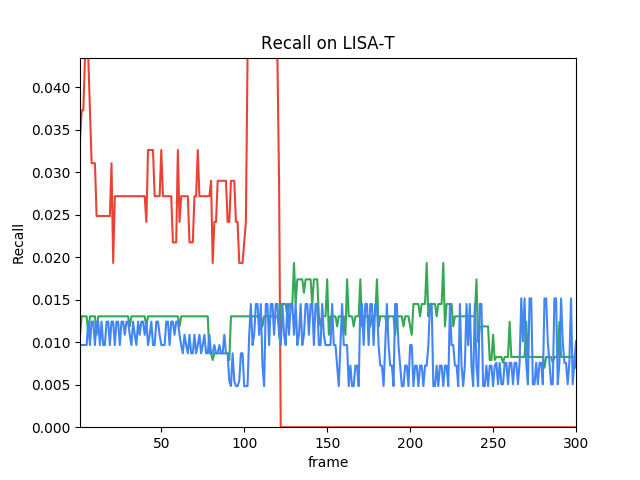}
    \caption{Recall}
    \label{fig:recall_lisa_t}
  \end{subfigure}
  \begin{subfigure}[t]{0.49\linewidth}
    \includegraphics[width=1.0\textwidth]{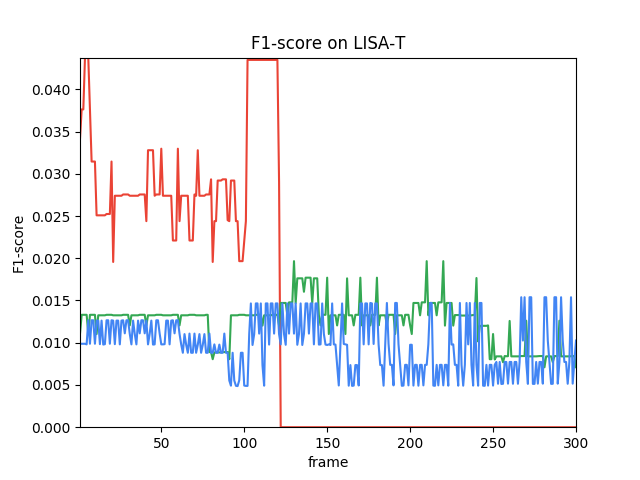}
    \caption{$F_1$-score}
    \label{fig:f1_score_lisa_t}
  \end{subfigure}
  \caption{Evaluation metrics on the \texttt{LISA-T} dataset. \textit{User1} annotated the objects very accurately in the beginning. \newline \protect\tikz \protect\fill[myred] (1ex,1ex) circle (1ex); User1 \qquad \protect\tikz \protect\fill[mygreen] (1ex,1ex) circle (1ex); User2 \qquad \protect\tikz \protect\fill[myblue] (1ex,1ex) circle (1ex); User3}
  \label{fig:metrics_lisa_t}
\end{figure}

Fig. \ref{fig:metrics_lisa_t} shows evaluation metrics (3D-IoU, precision, recall and $F_1$-score) for the annotated \texttt{LISA-T} dataset. We provide the mathematical descriptions of each of these metrics below:

\begin{equation}
    precision = \frac{TP}{TP+FP},
\end{equation}

\begin{equation}
    recall = \frac{TP}{TP+FN},
\end{equation}

\begin{equation}
    F_1-score = \frac{1}{\frac{\frac{1}{recall}+(\frac{1}{precision}}{2}}.  
\end{equation}
Precision and Recall are standard metrics that evaluate the overall performance of statistical models, and the $F_1$-score calculates the harmonic mean of the precision and recall (harmonic mean because the precision and recall are ratios) \cite{goutteProbabilisticInterpretationPrecision2005}.

After the experiment, all participants were asked to fill out a feedback form to evaluate the performance and user experience with the annotation toolbox. Table \ref{tbl:feedback} shows the average results. From these results, we gather that the annotators described the toolbox as efficient, helpful, intelligent and easy to learn. The average score is 4.12 out of 5.00 which is 82.40\%. 

\begin{table}[t]
  \centering
  \caption{Evaluation of the user experience. All annotators rated the \textit{Masterview} with the highest score because it simplified the annotation task a lot.}
  \label{tbl:feedback}
  \begin{tabular}
    {|p{7.15cm}|r|}
    \hline
    \scriptsize Criteria                           & \scriptsize Score\\
    \hline\hline
     Does this toolbox increase your productivity?& 4.33/5  \\
     How would you rate the performance of the annotation tool?& 3.67/5  \\
     Is the annotation toolbox useful for you? & 4.50/5\\
     Is the toolbox efficient? & 4.00/5\\
     Is the interaction with the toolbox clear and understandable? & 3.67/5\\
     Are screen message consistent? &3.67/5\\
     Are input prompts clear?&4.00/5\\
     Are error messages helpful?&3.67/5\\
     Was it easy to learn to operate the annotation toolbox?&4.00/5\\
     How easy was the exploration of new features?&3.67/5\\
     Was it easy to remember all commands?&4.33/5\\
     Were help messages on the screen helpful?&3.67/5\\
     Was the supplemental reference material clear?&3.67/5\\
     Was it easy to read and understand the instructions?&4.33/5\\
     \textbf{The Masterview (side view, front view and bird's-eye-view) simplifies the task.}&\textbf{5.00/5}\\
     The organization of information is very clear.&4.00/5\\
     The system speed is fast enough to run the toolbox.&3.67/5\\
     The projection of 3D labels into the image domain is useful.&4.33/5\\
     The interpolation mode helps me to annotate more labels in the same amount of time.&4.67/5\\
     The auto-save functionality was helpful to save time.&4.67/5\\
     The undo/redo functionality follows an important principle of user interface design and increases the satisfaction of users.&4.00/5\\
     \hline
     & \bf{4.12/5}\\
    \hline
  \end{tabular}
\end{table}

\begin{figure}[t]
  \centering
  \captionsetup{justification=centering}
  \begin{subfigure}[t]{0.49\linewidth}
    \includegraphics[width=1.0\textwidth]{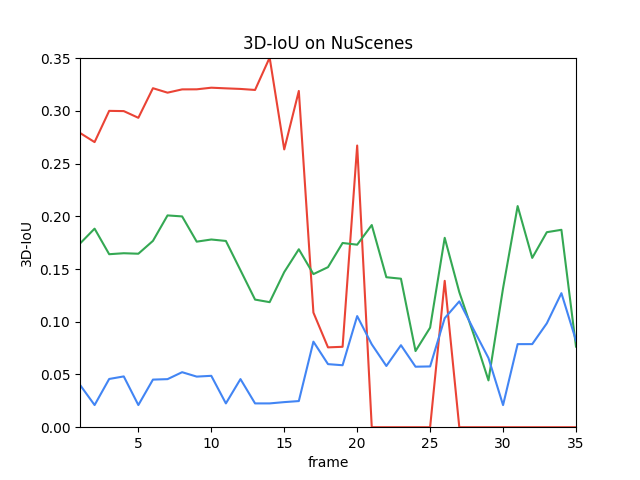}
    \caption{3D-IoU}
    \label{fig:3d_iou_nuscenes}
  \end{subfigure}
  \begin{subfigure}[t]{0.49\linewidth}
    \includegraphics[width=1.0\textwidth]{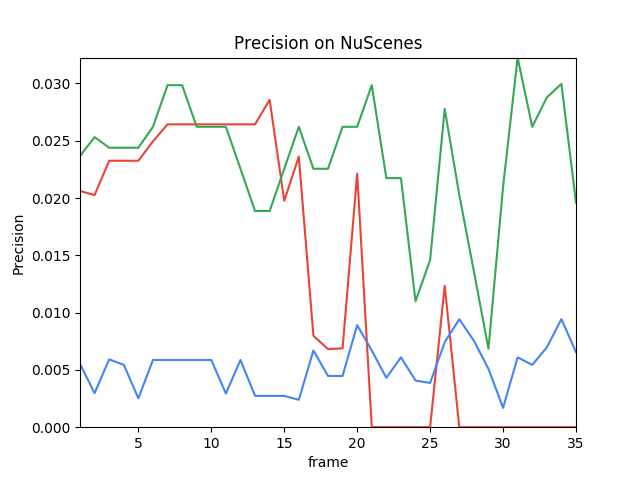}
    \caption{Precision}
    \label{fig:precision_nuscenes}
  \end{subfigure}
  \begin{subfigure}[t]{0.49\linewidth}
    \includegraphics[width=1.0\textwidth]{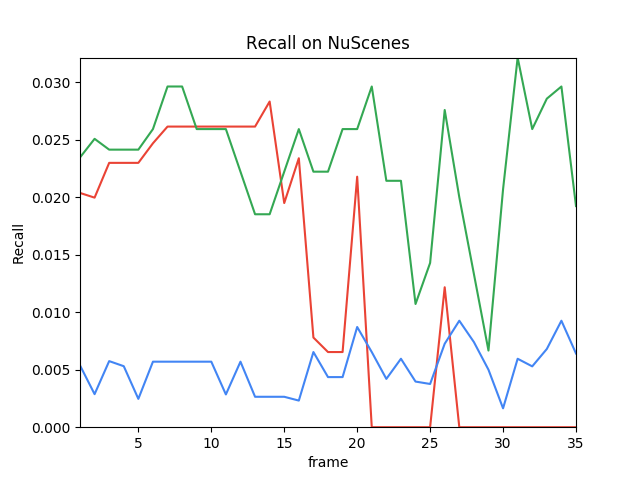}
    \caption{Recall}
    \label{fig:recall_nuscenes}
  \end{subfigure}
  \begin{subfigure}[t]{0.49\linewidth}
    \includegraphics[width=1.0\textwidth]{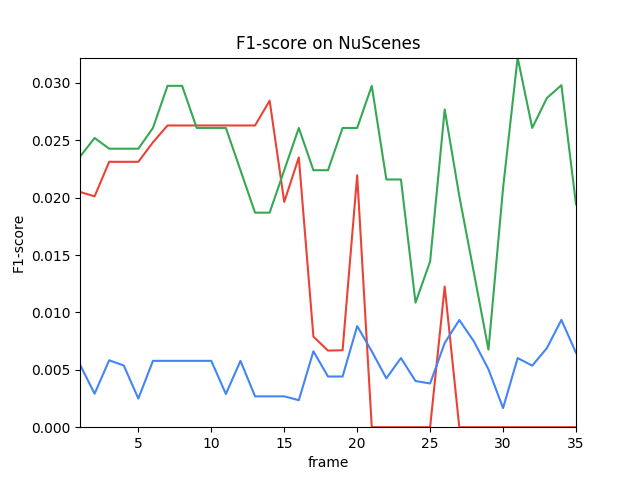}
    \caption{$F_1$-score}
    \label{fig:f1_score_nuscenes}
  \end{subfigure}
  \caption{Evaluation metrics on the \textit{NuScenes} dataset. \textit{User2} achieved the best accuracy in total. \newline \protect\tikz \protect\fill[myred] (1ex,1ex) circle (1ex); User1 \qquad \protect\tikz \protect\fill[mygreen] (1ex,1ex) circle (1ex); User2 \qquad \protect\tikz \protect\fill[myblue] (1ex,1ex) circle (1ex); User3}
  \label{fig:metrics_nuscenes}
\end{figure}

Evaluation was also performed on the \texttt{NuScenes} dataset since groundtruth annotations were already provided. 
In the second experiment again user statistics were created (see Table \ref{tbl:statistics_nuscenes}).
The average 3D-IoU between the manually created ground truth and the user annotations is 0.0748 and 0.0393 (considering the boundary of 0.6). 
Fig. \ref{fig:metrics_nuscenes} shows evaluation metrics (3D-IoU, precision, recall and $F_1$-score) for each annotated frame on the first \texttt{NuScenes} sequence.

\begin{table}[t]
  \centering
  \caption{Annotation statistics of different users on the \texttt{NuScenes} dataset. Annotators who spent more time on the annotation task achieved better accuracy.}
  \label{tbl:statistics_nuscenes}
  \begin{tabular}{|l|llll|}
    \hline
                             & Ground truth&User1    & User2    & User3\\
                             \hline\hline
    Time (hh:mm):         &  04:00  & \textbf{00:49} & 02:30 & 01:30 \\
    Clicks:               &  4933      & 2815     & 4418     & \textbf{2440} \\
    Keystrokes:           & 5812       & 1521     & 1488     & \textbf{134} \\
    Annotated obj.:       & \textbf{1303}       & 388      & 395      & 585\\
    \quad CAR:            & \textbf{1048}       & 374      & 395      & 585\\
    \quad PEDESTRIAN:     & \textbf{97}         & 14       & 0        & 0\\
    \quad MOTORCYCLE:     & 0          & 0        & 0        & 0\\
    \quad BICYCLE:        & 0          & 0        & 0        & 0\\
    \quad TRUCK:          & \textbf{158}        & 0        & 0        & 0\\
    3D-IoU:               & -          & 0.068  & \textbf{0.084}  & 0.072\\
    3D-IoU\textgreater0.6:& -          & 0.047  & \textbf{0.054}  & 0.017\\
    Precision:            &  -         & 0.004  & \textbf{0.015}  & 0.002\\
    Recall:               &  -         & 0.004  & \textbf{0.014}  & 0.002\\
    $F_1$ score:          &  -         & 0.004  & \textbf{0.015}  & 0.002\\
    \hline
  \end{tabular}
\end{table}



\begin{figure*}[t]
  \centering
  \includegraphics[width=0.95\textwidth, height=2in]{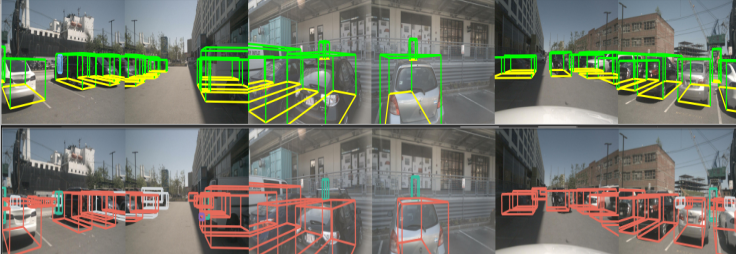}
  \caption{Comparison between manually created annotations (top) and annotations provided by \texttt{NuScenes} (bottom).}
  \label{fig:comparison_with_nuscenes_annotations}
\end{figure*}

Next, we focus on the qualitative evaluation of our annotation tool. Fig.~\ref{fig:comparison_with_nuscenes_annotations} shows a comparison between manually annotated objects (top) and annotations provided by \texttt{NuScenes} (bottom).
One can see that \texttt{NuScenes} provides a lot of annotations. Even objects that can't be seen in the camera image were annotated. The reason for that is that objects containing at least one point of the point clouds were annotated \cite{caesarDevkitNuScenesDataset2019}.

\section{Concluding Remarks \& Future Directions}
In this research we have contributed to three specific areas. First, we proposed a novel annotation system with tools for efficient, accurate 3D localization of objects and their dynamic movement using full-surround multi-modal (camera and LiDAR) data streams. A robust, scalable and efficient way to annotate 2D and 3D data was provided. The main goal was to create an annotation tool that enables the user to obtain full-surround annotations as well as annotations of tracks. Second, a systematic comparison with 33 annotation tools was performed. Finally, an evaluation of the efficiency and accuracy of the system with four human annotators with diverse backgrounds and skills was carried out, resulting in more than 13,800 object annotations within one hour.

The next immediate step is to test the annotation toolbox with multiple users, to perform stress tests and make the toolbox ready to be used by multiple users in parallel. A server-side application will be implemented that is connected to a database to obtain and store labels from the crowd. 

The second goal is to use a small amount of the dataset that was manually annotated to train object detection algorithms to provide an inference feature that predicts the pose of objects. Using this feature, annotators only need to perform small adjustments and sometimes delete false positives. Annotators also mentioned that interpolation for multiple objects would be very helpful which would further decrease annotation time.

Next, a point cloud aggregation mode will be added to facilitate the labeling of distant objects. Also, a variant of the Iterative Closest Point (ICP) algorithm will be used to ensure that labels move in the correct direction when the user switches to the next frame. Moreover, further annotation modes (attribute annotation, key point annotation, semantic annotation as well as behavior annotation) will be added to the toolbox. Finally, a highly automatic annotation mode will enable the user to obtain annotations very efficiently so that only small corrections by the human are needed.






\bibliographystyle{IEEEtran}
\bibliography{references}

\end{document}